\documentclass{article}

\usepackage[final]{corl_2026}
\usepackage[utf8]{inputenc}
\usepackage[T1]{fontenc}
\usepackage{url}
\usepackage{nicefrac}
\usepackage[table]{xcolor}
\usepackage{tabularx}
\usepackage{microtype}
\usepackage{graphicx}
\usepackage{subcaption}
\usepackage{booktabs}

\usepackage{hyperref}

\usepackage{amsmath}
\usepackage{amssymb}
\usepackage{mathtools}
\usepackage{amsthm}
\usepackage{multirow}
\usepackage{makecell}
\usepackage{xspace}
\usepackage{algorithm}
\usepackage{algpseudocode}
\usepackage{cleveref}
\crefname{figure}{fig.}{figs.}
\Crefname{figure}{Fig.}{Figs.}
\usepackage{wrapfig}
\usepackage{marvosym}

\newcommand{\ours}[0]{{HAF-VLA}\xspace}
\title{HAF-VLA: Hierarchical Action Flow Framework for
Humanoid Vision-Language-Action Control
}

\author{
\begin{tabular}{@{}p{\linewidth}@{}}
\centering
Langzhe Gu\textsuperscript{1,2,*},
\textbf{Chengkai Hou}\textsuperscript{1,2,*},
\textbf{Meng Li}\textsuperscript{2,*},
\textbf{Xinhua Wang}\textsuperscript{2},
\textbf{Xinyuan Lv}\textsuperscript{2,3},\\
\textbf{Shuanghao Bai}\textsuperscript{2,4},
\textbf{Guangrun Li}\textsuperscript{1,2},
\textbf{Jingyang He}\textsuperscript{1,2},
\textbf{Ziluo Ding}\textsuperscript{2},
\textbf{Zhiyuan Xu}\textsuperscript{2},\\
\textbf{Ning Ding}\textsuperscript{5},
\textbf{Kuan Cheng}\textsuperscript{1},
\textbf{Jian Tang}\textsuperscript{2,\Letter},
\textbf{Shanghang Zhang}\textsuperscript{1,\Letter},
\textbf{Zhengping Che}\textsuperscript{2,\Letter}
\\[0.5em]
\textsuperscript{1}State Key Laboratory of Multimedia Information Processing, School of Computer Science, Peking University\\
\textsuperscript{2}Beijing Innovation Center of Humanoid Robotics\\
\textsuperscript{3}Nankai University\quad \textsuperscript{4}Xi’an Jiaotong University\quad \textsuperscript{5}Tsinghua University
\\[0.3em]
\textsuperscript{*}Equal contribution.
\qquad
\textsuperscript{\Letter}Corresponding author.
\end{tabular}
}

\hypersetup{
    pdftitle={HAF-VLA: Hierarchical Action Flow Framework
              for Humanoid Vision-Language-Action Control},
    pdfauthor={Langzhe Gu et al.}
}

\begin{document}
\maketitle

\begin{abstract}
    Humanoid robots hold great promise as general-purpose agents in human-centered environments, yet adapting generalist vision-language-action (VLA) policies to real-world humanoid loco-manipulation remains challenging. Unlike fixed-base or wheeled manipulators, humanoids must coordinate locomotion, posture, gaze, and dual-arm manipulation over a high-dimensional whole-body action space, where inaccurate base or waist motions can easily induce unstable and aggressive upper-body compensation. We propose HAF-VLA, a hierarchical action-flow framework for humanoid VLA control. Built upon a pretrained flow-matching VLA, HAF-VLA inherits the vision-language prefix encoder and action expert weights, and adapts them to humanoid whole-body loco-manipulation with learnable stage embeddings and cross-stage KV-cache conditioning. Instead of generating the full action chunk in a single pass, HAF-VLA progressively denoises a shared whole-body action space: first locomotion and head commands, then waist refinement, and finally full-body manipulation. To preserve structural dependencies across stages, we introduce a cross-stage KV cache conditioning mechanism, enabling subsequent action generation stages to refine and augment their outputs based on the action representations from preceding stages. Experimental results across six real-world tasks demonstrate that our framework significantly enhances the performance of existing models in humanoid loco-manipulation tasks, demonstrating an effective pathway for adapting strong mobile-manipulation VLAs to humanoid whole-body loco-manipulation.
\end{abstract}


\begin{figure}[htbp]
  \centering
  \includegraphics[width=1.0\textwidth]{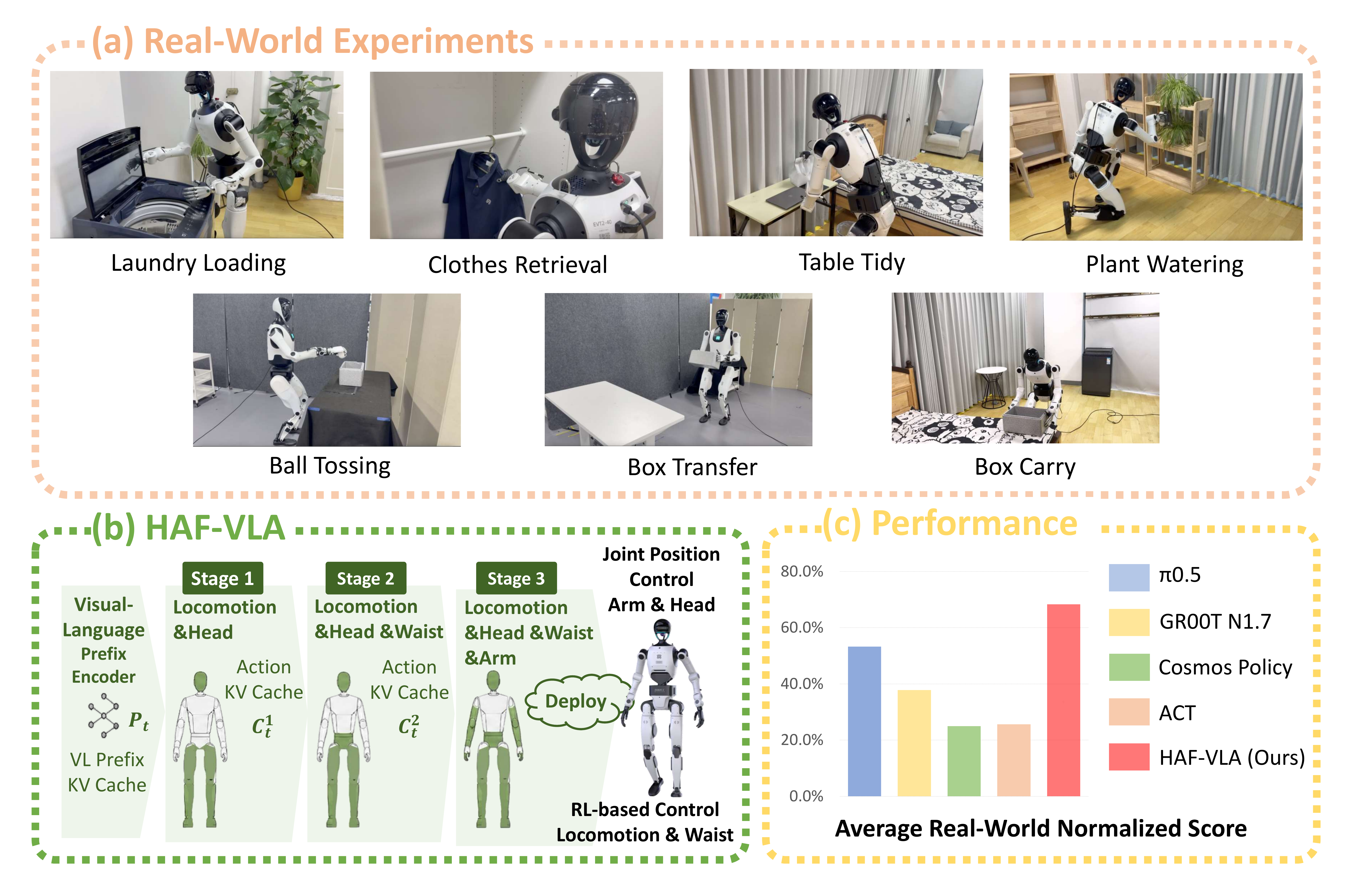}
  \caption{\textbf{Overview of \ours.}  (a) We demonstrate seven loco-manipulation tasks tested on the
humanoid robot in real-world scenarios. (b) The pipeline illustrating our hierarchical approach
to generating actions across different robot stages. (c) Average Real-World Normalized Score of
different baselines and HAF-VLA on the six real-world tasks. }
  \label{fig:method_overview}
  \vspace{-2em}
\end{figure}

\section{Introduction}

Humanoid robots hold the promise of becoming general-purpose agents capable of operating in human-centered environments. Unlike fixed-base or wheeled robots, humanoids must coordinate locomotion, waist motion, gaze, and dual-arm manipulation to navigate large spaces, adjust posture, maintain balance, and physically interact with the environment~\cite{homie,falcon,softa,almi,exbody2,liao2025beyondmimicmotiontrackingversatile,ze2025generalizable,jiang2026wholebodyvla}. This whole-body nature makes humanoid loco-manipulation fundamentally different from conventional manipulation: inaccurate base or waist motions can force the upper body to produce compensatory movements, degrading precision, stability, and energy efficiency.

Recent vision-language-action models and world-action models have achieved strong results on fixed-base, dual-arm, and wheeled mobile manipulators~\cite{pi0, intelligence2025pi05visionlanguageactionmodelopenworld, gr00tn1_2025, kim2026cosmos, openvlaoft, fan2025xr}. However, most of them are not explicitly designed for the structured action space and unstable bipedal dynamics of humanoids. Directly applying a single-stage action generator to humanoid loco-manipulation may entangle locomotion, posture, and manipulation in an unstructured manner, making the policy prone to unstable base motions or aggressive upper-body compensation~\cite{exbody2, liao2025beyondmimicmotiontrackingversatile,omnih2o,r2s2}. Therefore, humanoid VLA control requires an action-generation structure that respects the kinematic dependencies among the legs, waist, head, and arms while preserving the semantic reasoning ability of pretrained VLA models.

To this end, we propose HAF-VLA, a hierarchical action-flow framework for humanoid vision-language-action control. Built upon a pretrained flow-matching VLA~\cite{intelligence2025pi05visionlanguageactionmodelopenworld}, HAF-VLA reuses its vision-language prefix encoder~\cite{beyer2024paligemmaversatile3bvlm} and action expert weights, and adapts them to humanoid whole-body loco-manipulation through progressive action generation. Instead of synthesizing the full action chunk in a single pass, HAF-VLA first generates locomotion and head commands, then incorporates waist refinement, and finally produces full-body manipulation actions. This design encourages locomotion-related motions to be planned before manipulation, reducing the tendency of the upper body to compensate for inaccurate base or waist movements~\cite{chen2025acdit, jiang2025behaviorrobotsuitestreamlining}.

We further introduce cross-stage KV-cache~\cite{liang2025mixtureoftransformerssparsescalablearchitecture} conditioning to propagate structured action information across the hierarchy (see \Cref{fig:method_overview}(b)). After each non-final stage completes flow denoising~\cite{yaron2023flowmatching}, its clean action prediction is re-encoded by the shared action expert to form an action KV cache, which conditions the following stage. In this way, later stages refine and augment their actions based on earlier motion plans, while all stages remain grounded in the same visual-language prefix representation.

We evaluate HAF-VLA on two full-sized humanoid platforms, Tian Gong 3.0 and Tian Gong 2.0, across six real-world loco-manipulation tasks spanning household and industrial scenarios (see \Cref{fig:method_overview}(a)). HAF-VLA consistently improves normalized task scores over imitation-learning, VLA, and world-model baselines  (see \Cref{fig:method_overview}(c)), especially on tasks requiring long-distance locomotion, precise body positioning, and coordinated whole-body manipulation. Our contributions are: (i) a hierarchical action-flow formulation for humanoid VLA loco-manipulation over a shared whole-body action space; (ii) a cross-stage KV-cache conditioning mechanism for progressive whole-body action generation; and (iii) real-world validation showing that this structure effectively adapts pretrained mobile-manipulation VLAs to humanoid loco-manipulation.

\section{Related Work}
\textbf{Vision-language-action models.}
Vision-language-action models (VLAs) aim to map visual observations and language instructions directly to robot actions~\cite{rdt, kim2025openvla, zheng2026xvla, univla}. Representative works such as RT-2~\cite{zitkovich2023rt2} and OpenVLA~\cite{kim2025openvla} show that vision-language pretraining can be effectively transferred to robotic manipulation, while $\pi_0$ and $\pi_{0.5}$ further move toward continuous generative action modeling with flow matching~\cite{pi0, intelligence2025pi05visionlanguageactionmodelopenworld}. However, most existing VLAs still focus on tabletop tasks or mobile manipulators, leaving high-dimensional humanoid whole-body action generation underexplored.

\textbf{Humanoid VLAs.}
Recent works have begun to extend VLAs to humanoid robots~\cite{shao2025langwbclanguagedirectedhumanoidwholebody, xue2025leverb, humanoidvla, bai2026hexhumanoidalignedexpertscrossembodiment}. GR00T N1.7~\cite{gr00tn1_2025} studies general-purpose humanoid action generation from vision-language inputs, while WholeBodyVLA~\cite{jiang2026wholebodyvla} explores latent VLA learning for large-space humanoid mobile manipulation. Recently, $\Psi_0$~\cite{wei2026psi0openfoundationmodel} introduced an open humanoid foundation model pretrained from egocentric human videos. In contrast to methods that rely on latent skills, cross-embodiment transfer, or learning from human videos, our method is directly built upon a pretrained generalist VLA and improves its performance on humanoid by generating structured whole-body actions grounded in the humanoid forward-kinematic chain, sparing the computational cost of training a new model.

\textbf{Whole-body manipulation.}
Recent advances in humanoid whole-body control have greatly expanded the motion repertoire of legged humanoids~\cite{liao2025beyondmimicmotiontrackingversatile, luo2025sonic, li2025bfmzeropromptablebehavioralfoundation, peng2018deepmimic}. With improved reinforcement learning and low-level control techniques, humanoid robots can now perform highly dynamic behaviors. Nevertheless, extending these capabilities to generalizable loco-manipulation remains substantially more challenging. Works such as TWIST2~\cite{ze2025twist2scalableportableholistic} and AMO~\cite{li2025amoadaptivemotionoptimization} demonstrate the potential of VR-based teleoperation for whole-body humanoid control and data collection. However, their focus is mainly on teleoperation interfaces and low-level control pipelines, rather than learning a generalizable policy for long-horizon loco-manipulation.

\vspace{-0.5em}

\section{\ours}
\begin{figure}[htbp]
\centering
\includegraphics[width=1.0\textwidth]{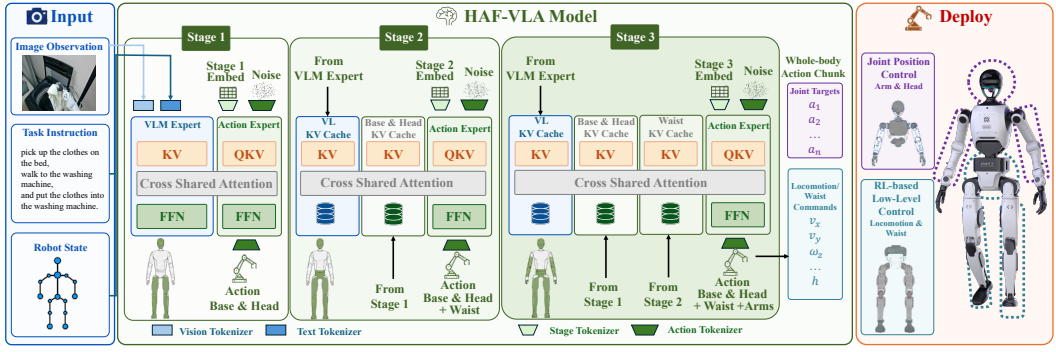}
\caption{\textbf{Overall framework of \ours.} Given visual observations, robot states, and language instructions, the model progressively generates whole-body action trajectories through hierarchical action-flow stages. The final action chunk is executed by the robot-specific low-level control interface.}
\label{fig:pipeline}
\end{figure}

\ours is built as a structured adaptation of a pretrained action-flow VLA. Specifically, we build upon $\pi_{0.5}$~\cite{intelligence2025pi05visionlanguageactionmodelopenworld}, inheriting its pretrained vision-language prefix encoder and action expert weights. We further introduce learnable stage embeddings and cross-stage KV-cache conditioning, transforming the original single-stage action generator into a hierarchical whole-body action-flow model. This design provides a simple and effective way to adapt a strong generalist mobile-manipulation VLA to high-dimensional humanoid whole-body loco-manipulation without training the full model from scratch. The detailed framework for \ours is illustrated in \Cref{fig:pipeline}. 

\subsection{Humanoid Action Interface}

Given a language instruction $\ell$ and the current observation $o_t$, \ours predicts a future whole-body action chunk \begin{equation} A_t = a_{t:t+H-1} \in \mathbb{R}^{H \times D}, \end{equation} where $o_t$ contains the egocentric RGB image $I_t$ and the proprioceptive robot state $q_t$. In our implementation, the action horizon is $H=100$
and the model action dimension is $D=32$,
comprising 26 effective control dimensions and
6 zero-padded dimensions. We define the effective humanoid action at each timestep as \begin{equation} a_t^{ctrl} = \left\{ q_{\mathrm{arm}}, q_{\mathrm{hand}}, v_x, v_y, \omega_z, q_{\mathrm{head}},w_{\mathrm{rpyz}},  c_{\mathrm{skill}} \right\}, \end{equation} where $q_{\mathrm{arm}}\in\mathbb{R}^{14}$ denotes dual-arm joint targets, $q_{\mathrm{hand}}\in\mathbb{R}^{2}$ denotes hand commands, $(v_x,v_y,\omega_z)\in\mathbb{R}^{3}$ denotes locomotion velocity commands, $w_{\mathrm{rpyz}}\in\mathbb{R}^{4}$ denotes waist roll, pitch, yaw, and height commands, $q_{\mathrm{head}}\in\mathbb{R}^{2}$ denotes head joint targets, and $c_{\mathrm{skill}}\in\mathbb{R}$ denotes the skill-mode command. The full model action $a_t$ is obtained by zero-padding
$a_t^{\mathrm{ctrl}}$ to 32 dimensions. The locomotion and waist commands are executed by existing RL-based low-level controllers~\cite{peng2018deepmimic, amp}, while the arm, hand, and head commands are sent to position-control interfaces. Robot-specific command conversion details are provided in the supplementary material.

\subsection{Hierarchical Action Flow}

Instead of generating the complete action chunk in the full action space at once, \ours decomposes action generation into three cumulative stages. Let the effective action dimension set be
\begin{equation}
\begin{aligned}
    \mathcal{D}
    =
    \mathcal{D}_{\mathrm{move}}
    \cup
    \mathcal{D}_{\mathrm{skill}}
    \cup
    \mathcal{D}_{\mathrm{head}}
    \cup
    \mathcal{D}_{\mathrm{waist}}
    \cup
    \mathcal{D}_{\mathrm{arm}}
    \cup
    \mathcal{D}_{\mathrm{hand}} .
\end{aligned}
\end{equation}
We define the active action dimensions of the three stages as
\begin{equation}
\begin{aligned}
    \mathcal{D}_1
    &=
    \mathcal{D}_{\mathrm{move}}
    \cup
    \mathcal{D}_{\mathrm{skill}}
    \cup
    \mathcal{D}_{\mathrm{head}}, \\
    \mathcal{D}_2
    &=
    \mathcal{D}_1
    \cup
    \mathcal{D}_{\mathrm{waist}}, \\
    \mathcal{D}_3
    &=
    \mathcal{D}_2
    \cup
    \mathcal{D}_{\mathrm{arm}}
    \cup
    \mathcal{D}_{\mathrm{hand}}
    =
    \mathcal{D}.
\end{aligned}
\end{equation}
Stage 1 predicts locomotion, skill mode, and head commands, establishing the coarse motion and gaze direction. Stage 2 additionally activates waist commands, allowing the model to refine the body posture and upper-body workspace conditioned on the Stage 1 motion plan. Stage 3 activates the full action space, including arms and hands, and produces the final whole-body action chunk for execution. This progressive design encourages the model to first stabilize locomotion-related motions before generating manipulation actions.

\subsection{Cross-Stage KV-Cache Conditioning}

\ours uses a shared action expert across all stages. Rather than assigning separate Transformer experts to different body parts, we condition the same action expert with a learnable stage embedding $e_s$, where $s\in{1,2,3}$. The stage embedding informs the action expert of the currently activated action subspace.

The visual-language prefix encoder $ f_{\mathrm{prefix}}$ is computed once and reused by all stages:
\begin{equation}
C_{\mathrm{vl}} = f_{\mathrm{prefix}}(o_t, \ell),
\end{equation}
where $C_{\mathrm{vl}}$ denotes the KV cache produced by the pretrained vision-language prefix encoder $ f_{\mathrm{prefix}}$. To propagate structured motion information across stages, Stage $s$ is conditioned on both $C_{\mathrm{vl}}$ and the action caches from preceding stages,
\begin{equation}
C_{<s} = {C_1,\ldots,C_{s-1}}.
\end{equation}

A key design is that the cross-stage action cache is generated from the completed clean action prediction of the previous stage, rather than from its intermediate noisy action input. Specifically, after a non-final stage $s\in{1,2}$ completes its flow denoising process and obtains a clean stage action prediction $\hat{A}^s$, we feed $\hat{A}^s$ through the shared action expert $f_{\mathrm{action}}$ again with the corresponding stage embedding and collect the resulting self-attention KV cache:
\begin{equation}
C_s =
f_{\mathrm{action}}
\left(
\hat{A}^s,
C_{\mathrm{vl}},
C_{<s},
e_s
\right),
\quad s\in{1,2}.
\end{equation}
Therefore, the cache represents the structured action plan produced by the completed previous stage, which then serves as conditioning context for the next stage. The final executed action chunk is always taken from Stage 3, while Stage 1 and Stage 2 only provide intermediate motion representations.

\subsection{Training and Receding-Horizon Inference}

\ours is trained with a stage-wise flow-matching objective. Given a clean action chunk $A_t\in\mathbb{R}^{H\times D}$, we sample Gaussian noise $\epsilon\sim\mathcal{N}(0,I)$ and a flow time $\tau\in[0,1]$. For each stage $s$, we define a binary mask $m_s\in\{0,1\}^{D}$, where $m_s^{(d)}=1$ if the $d$-th action dimension belongs to $\mathcal{D}_s$ and $m_s^{(d)}=0$ otherwise. The mask is broadcast along the temporal dimension, allowing all stages to operate in the same global action space while enforcing zero target velocities on inactive dimensions.

For each stage $s$, the noisy action input and
target velocity are masked as
\begin{equation}
X_\tau^s =
\big((1-\tau)\epsilon+\tau A_t\big)\odot m_s,
\qquad
u^s=(A_t-\epsilon)\odot m_s.
\end{equation}
The action expert predicts the velocity field
$\hat{u}^s$ over the full action space.
We compute the flow-matching loss over all action
dimensions, including inactive dimensions whose
target velocities are zero:
\begin{equation}
\mathcal{L}_{\mathrm{FM}}
=
\sum_{s=1}^{3}
\frac{1}{HD}
\left\|\hat{u}^s-u^s\right\|_F^2.
\end{equation}

During training, we use teacher forcing for cross-stage cache construction. For non-final stages, the cache used by the following stage is generated from the ground-truth masked action chunk rather than from the model prediction. This stabilizes optimization while preserving the same sequential dependency structure used at inference time. During inference, the model executes the three stages sequentially. Each stage starts from independently sampled noise and performs 10 flow-denoising steps conditioned on the visual-language cache and the available cross-stage action caches.

We deploy \ours with a receding-horizon execution scheme. The robot executes the first 40 steps before the next inference call. In our real-robot deployment, the complete three-stage inference takes approximately 0.12 seconds on a single RTX 5090 GPU, supporting real-time closed-loop humanoid loco-manipulation.

\section{Experiments}
\label{sec:result}

We evaluate \ours through three questions:
\textbf{Q1.} Does \ours enable long-horizon humanoid loco--manipulation beyond existing SOTA methods?
\textbf{Q2.} Which design choices contribute to the performance of \ours?
\textbf{Q3.} Does \ours generalize to unseen distribution shifts?

\subsection{Experiment Setup}
\label{sec:experiment_setup}

\begin{wrapfigure}{r}{0.6\textwidth}
  \centering
  \includegraphics[width=0.6\textwidth]{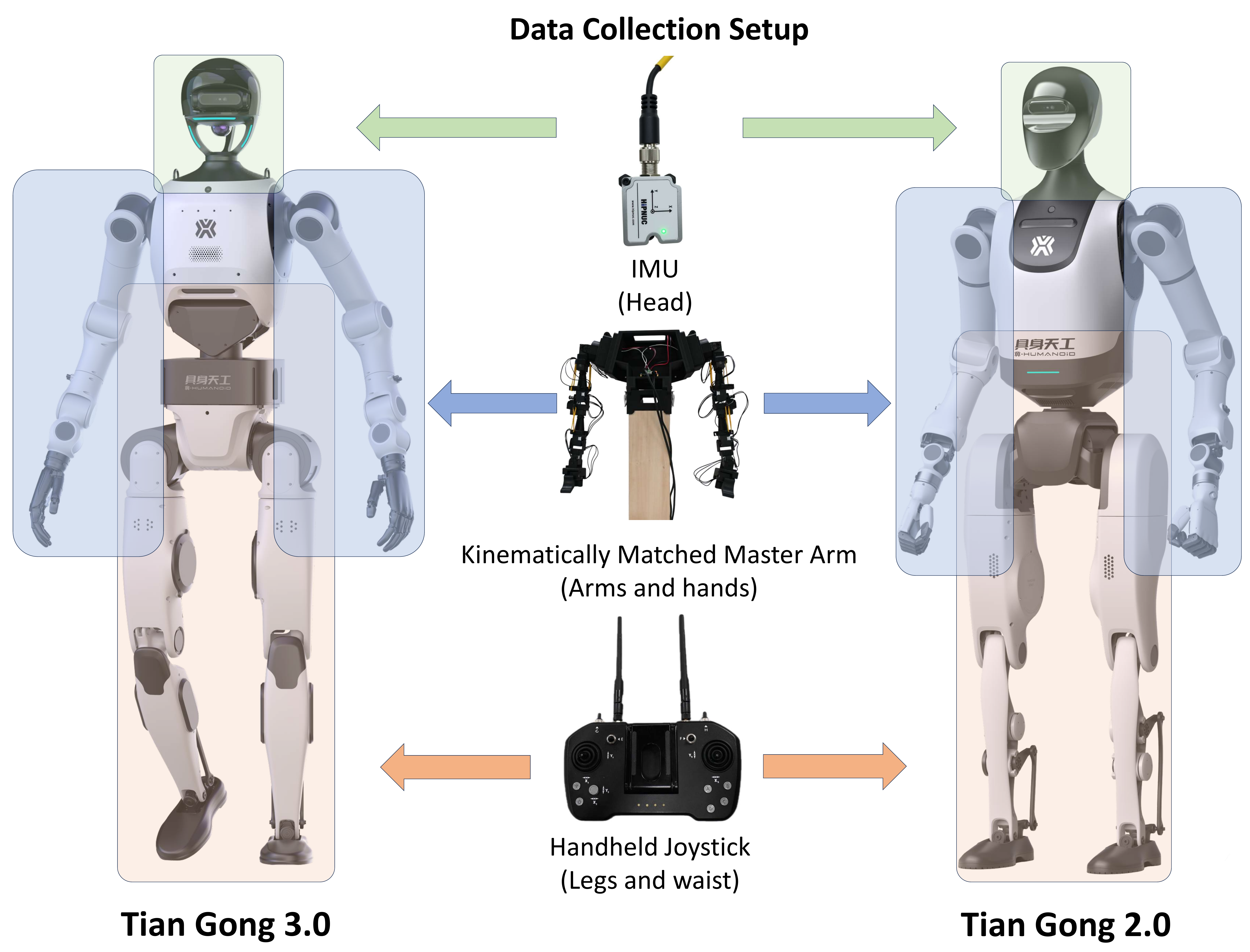}
  \caption{
  \textbf{Robot teleoperation data collection setup.}
  An IMU controls the head, kinematically matched master arms control the arms and hands, and a handheld joystick controls locomotion and waist motion.
  }
  \label{fig:data_collection}
  \vspace{-1em}
\end{wrapfigure}

\noindent\textbf{Hardware and Data Collection.}
We conduct all experiments on the Tian Gong 3.0 and Tian Gong 2.0 humanoid robots using an onboard egocentric RGB camera.
Demonstrations are collected through isomorphic teleoperation~\cite{hou2026robomind20multimodalbimanual}: kinematically matched master arms control dual-arm manipulation, a handheld joystick controls locomotion and waist motion, and an IMU controls the head, as shown in \Cref{fig:data_collection}.
We collect 120 teleoperated demonstrations for each task.
Additional robot and data-collection details are provided in the supplementary material.

\begin{figure}[htbp]
  \centering
  \includegraphics[width=1.0\textwidth]{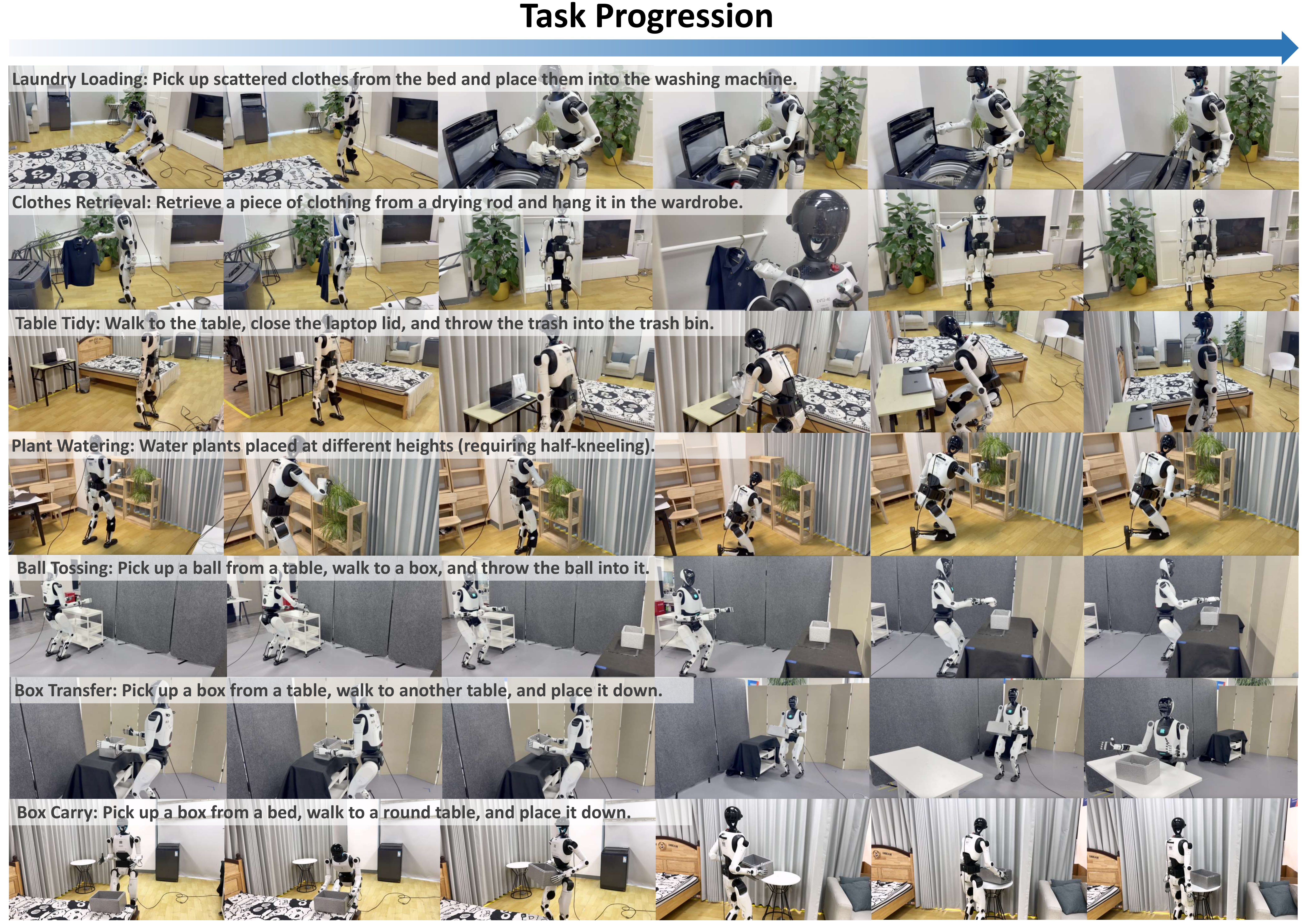}
  \caption{
  \textbf{Real-world evaluation tasks.}
  The first six tasks constitute the main benchmark used in Q1, while \textit{Box Carry} is used for the controlled ablation study in Q2.
  Each row shows a representative execution sequence requiring locomotion, whole-body posture adjustment, and object interaction.
  }
  \label{fig:task_progress}
  \vspace{-1em}
\end{figure}

\noindent\textbf{Tasks.}
As shown in \Cref{fig:task_progress}, we consider seven real-world humanoid loco-manipulation tasks requiring different combinations of locomotion, posture adjustment, dual-arm manipulation, and object interaction.
Six tasks---\textit{Laundry Loading}, \textit{Clothes Retrieval}, \textit{Table Tidy}, \textit{Plant Watering}, \textit{Ball Tossing}, and \textit{Box Transfer}---form the main benchmark used for comparison with existing methods in Q1.
We additionally use \textit{Box Carry}, in which the robot transports a box from a bed to a distant table, as a controlled ablation task in Q2.
The generalization experiments in Q3 introduce distribution shifts to \textit{Laundry Loading}, \textit{Clothes Retrieval}, and \textit{Box Carry}.

\noindent\textbf{Evaluation Metric.}
All quantitative evaluations use the same normalized task score.
Each task is decomposed into task-specific milestones with predefined scores, and the score achieved in each rollout $s_{\mathrm{achieved}}$ is normalized by the maximum achievable score $s_{\mathrm{max}}$:
$
\mathrm{Score}(\%) =
\frac{s_{\mathrm{achieved}}}{s_{\mathrm{max}}}\times100.
$
This metric captures partial progress in long-horizon tasks while allowing comparison across tasks with different completion criteria.
Unless otherwise specified, we report the average normalized score over 10 real-robot rollouts for each method and evaluation setting.
For variants that produce unsafe motions, evaluation is terminated and reported as ``Unsafe.''

\noindent\textbf{Baselines.}
For the main benchmark, we compare with four representative methods:
(i) ACT~\cite{act}, a classical imitation-learning policy;
(ii) $\pi_{0.5}$~\cite{intelligence2025pi05visionlanguageactionmodelopenworld}, a generalist flow-matching VLA;
(iii) GR00T N1.7~\cite{gr00tn1_2025}, a humanoid foundation model; and
(iv) Cosmos Policy~\cite{kim2026cosmos}, a world-model-based robotic policy.
We follow their official implementations and released configurations as closely as possible.
Detailed training settings and hyperparameters are provided in the supplementary material.
For controlled ablations and generalization experiments, we use $\pi_{0.5}$ as the primary single-stage VLA reference when applicable.

\vspace{-0.3em}
\subsection{Q1: Does \ours enable long-horizon humanoid loco--manipulation beyond existing SOTA methods?}
\label{sec:q1_main_exp}

We first evaluate \ours on the six main benchmark tasks in \Cref{fig:task_progress}.
As shown in \Cref{tab:vla_score}, \ours achieves the best or tied-best normalized score on all six tasks and improves the average score from 53.3\% for $\pi_{0.5}$, the strongest baseline on average, to 68.3\% (+15.0 percentage points).
The gains are particularly pronounced on \textit{Box Transfer}, \textit{Ball Tossing}, and \textit{Laundry Loading}, which require locomotion over extended spatial regions followed by precise manipulation.
We observe that $\pi_{0.5}$ and GR00T N1.7 are more prone to locomotion errors in these tasks, while Cosmos Policy incurs substantially higher inference latency in our implementation.
Overall, the results demonstrate the effectiveness of \ours for long-horizon humanoid whole-body loco-manipulation.

\begin{table*}[htbp]
    \centering
    \small
    \setlength{\tabcolsep}{5.5pt}
    \renewcommand{\arraystretch}{1.15}
    \caption{
    \textbf{Main results on six long-horizon humanoid loco--manipulation tasks.}
    We report the average normalized task score (\%) over 10 real-robot rollouts.
    Higher is better.
    }
    \label{tab:vla_score}
    \begin{tabular}{lccccc}
        \toprule
        \textbf{Task}
        & \textbf{HAF-VLA (Ours)}
        & $\boldsymbol{\pi_{0.5}}$
        & \textbf{GR00T N1.7}
        & \textbf{Cosmos Policy}
        & \textbf{ACT} \\
        \midrule
        Laundry Loading
        & \textbf{66.7} & 53.3 & 40.0 & 0.0 & 10.0 \\
        Clothes Retrieval
        & \textbf{53.3} & \textbf{53.3} & 33.3 & 26.7 & 23.3 \\
        Table Tidy
        & \textbf{80.0} & 70.0 & 40.0 & 16.7 & 23.3 \\
        Plant Watering
        & \textbf{60.0} & 50.0 & 33.3 & 30.0 & 16.7 \\
        Ball Tossing
        & \textbf{56.7} & 33.3 & 36.7 & 3.3 & 30.0 \\
        Box Transfer
        & \textbf{93.3} & 60.0 & 43.3 & 73.3 & 50.0 \\
        \midrule
        \textbf{Average}
        & \textbf{68.3} & 53.3 & 37.8 & 25.0 & 25.6 \\
        \bottomrule
    \end{tabular}
\end{table*}

\vspace{-0.6em}

\subsection{Q2: Which design choices contribute to the performance of \ours?}
\label{sec:q2_ablation}

We conduct controlled ablations on the \textit{Box Carry} task shown in \Cref{fig:task_progress}.
Unless otherwise specified, \ours uses three cumulative stages---locomotion/head, waist, and arms/hands---with 10 flow-matching denoising steps per stage.
All variants are trained on the same 120 demonstrations and evaluated using the normalized task score defined in \Cref{sec:experiment_setup}.
Results are summarized in \Cref{tab:ablation_study}.

\begin{table*}[htbp]
    \centering
    \small
    \setlength{\tabcolsep}{3.5pt}
    \renewcommand{\arraystretch}{1.10}
    \caption{
    \textbf{Ablation study of HAF-VLA on Box Carry.}
    We report the average normalized task score (\%) over 10 real-robot rollouts.
    ``Unsafe'' denotes variants terminated due to unsafe motions.
    }
    \label{tab:ablation_study}

    \begin{tabularx}{\textwidth}{
        @{}
        >{\raggedright\arraybackslash}p{2.2cm}
        >{\raggedright\arraybackslash}p{3.1cm}
        >{\raggedright\arraybackslash}X
        >{\centering\arraybackslash}p{1.25cm}
        @{}
    }
        \toprule
        \textbf{Component}
        & \textbf{Variant}
        & \textbf{Design}
        & \textbf{Score} \\
        \midrule

        \multirow{7}{*}{Stage design}
        & \textbf{HAF-VLA (Ours)}
        & Loco. $\rightarrow$ Waist $\rightarrow$ Arms
        & \textbf{86.7} \\

        & Waist-First
        & Waist $\rightarrow$ Arms $\rightarrow$ Loco.
        & 70.0 \\

        & Arm-First
        & Arms $\rightarrow$ Waist $\rightarrow$ Loco.
        & 60.0 \\

        & Loss-Calibrated
        & Reweight action groups by $1/3,\;1/2,\;1$
        & 76.7 \\

        & All-Joint
        & Denoise all joints at every stage
        & 73.3 \\

        & $\pi_{0.5}$ (10 steps)
        & Single-stage full-body denoising
        & 70.0 \\

        & $\pi_{0.5}$ (30 steps)
        & Single-stage full-body denoising
        & 30.0 \\
        \midrule

        \multirow{3}{*}{KV cache}
        & No-Cache-Train
        & Remove cache during training and deployment
        & 50.0 \\

        & No-Cache-Deploy
        & Remove cache only at deployment
        & Unsafe \\

        & Zero-Cache-Deploy
        & Zero cache only at deployment
        & Unsafe \\
        \midrule

        \multirow{2}{*}{Refinement}
        & Hard-Freeze
        & Freeze previously generated action dimensions
        & 66.7 \\

        & Stage-Concat
        & Concatenate stage-wise outputs without refinement
        & 46.7 \\

        \bottomrule
    \end{tabularx}
\end{table*}
\vspace{-0.5em}

\noindent\textbf{Stage design.}
We first examine the stage order by rearranging the action groups while preserving the three-stage structure.
The proposed locomotion$\rightarrow$waist$\rightarrow$arms hierarchy achieves 86.7\%, outperforming Waist-First (70.0\%) and Arm-First (60.0\%).
To account for unequal supervision frequencies across stages, \textit{Loss-Calibrated} reweights the losses of locomotion/head, waist, and arms/hands by $1/3$, $1/2$, and $1$, respectively, achieving 76.7\%.
These results support the proposed stage ordering beyond differences in supervision frequency.

We further evaluate generation without body-part-specific staging.
\textit{All-Joint}, which activates the full action space at every stage, achieves 73.3\%.
To control for the total denoising budget, we compare \ours with single-stage $\pi_{0.5}$ using 30 denoising steps.
Despite the same total number of steps, $\pi_{0.5}$ achieves only 30.0\%, compared with 86.7\% for \ours, demonstrating that the improvement cannot be attributed to additional denoising steps alone.
Notably, increasing $\pi_{0.5}$ from 10 to 30 steps reduces its score from 70.0\% to 30.0\%, potentially due in part to increased inference latency (0.075\,s to 0.115\,s).
In contrast, \ours maintains strong performance despite a comparable latency of 0.123\,s, highlighting the benefit of hierarchical action generation.

\noindent\textbf{KV-cache conditioning.}
We examine the role of cross-stage conditioning by removing the action KV cache.
\textit{No-Cache-Train}, which disables the cache during both training and deployment, reduces performance from 86.7\% to 50.0\%.
In contrast, \textit{No-Cache-Deploy} and \textit{Zero-Cache-Deploy} remove or zero the cache only at inference time, respectively, resulting in unsafe motions.
These results highlight the importance of cross-stage information propagation and consistent conditioning between training and deployment.

\begin{figure}[htbp!]
    \centering
    \includegraphics[width=0.8\columnwidth]{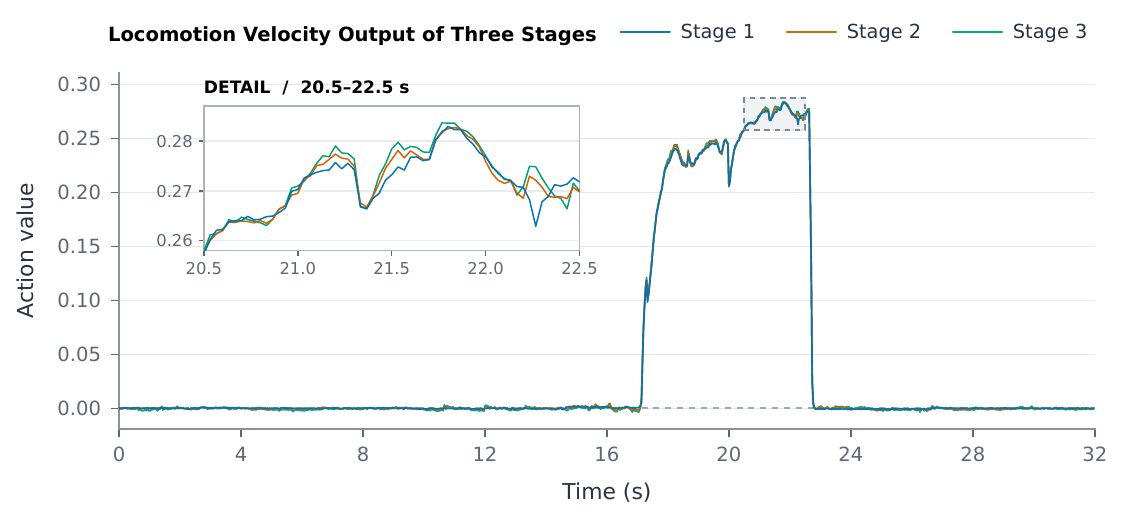}
    \caption{
    \textbf{Cross-stage consistency and refinement.}
    Forward-velocity predictions from three stages recorded during a real-world \textit{Box Carry} rollout.
    The stages maintain a consistent overall trajectory, while later stages make localized refinements.
    }
    \label{fig:vx_3stage}
    \vspace{-1.0em}
\end{figure}

\noindent\textbf{Cross-stage refinement.}
We evaluate whether later stages benefit from revising previously generated actions.
\textit{Hard-Freeze} prevents later stages from modifying earlier-stage action dimensions, while \textit{Stage-Concat} constructs the final action by concatenating the corresponding outputs from individual stages.
Both underperform the proposed soft refinement, achieving 66.7\% and 46.7\%, respectively.
\Cref{fig:vx_3stage} further visualizes the action outputs of all three stages recorded during a real-world \textit{Box Carry} rollout.
The predictions preserve a highly consistent overall locomotion trajectory, while later stages make small, localized adjustments to earlier-stage outputs.

\vspace{-0.5em}

\vspace{-0.3em}
\subsection{Q3: Does \ours generalize to unseen distribution shifts?}
\label{sec:q3_exp_generalization}

\begin{figure}[htbp!]
  \centering
  \includegraphics[width=0.9\textwidth]{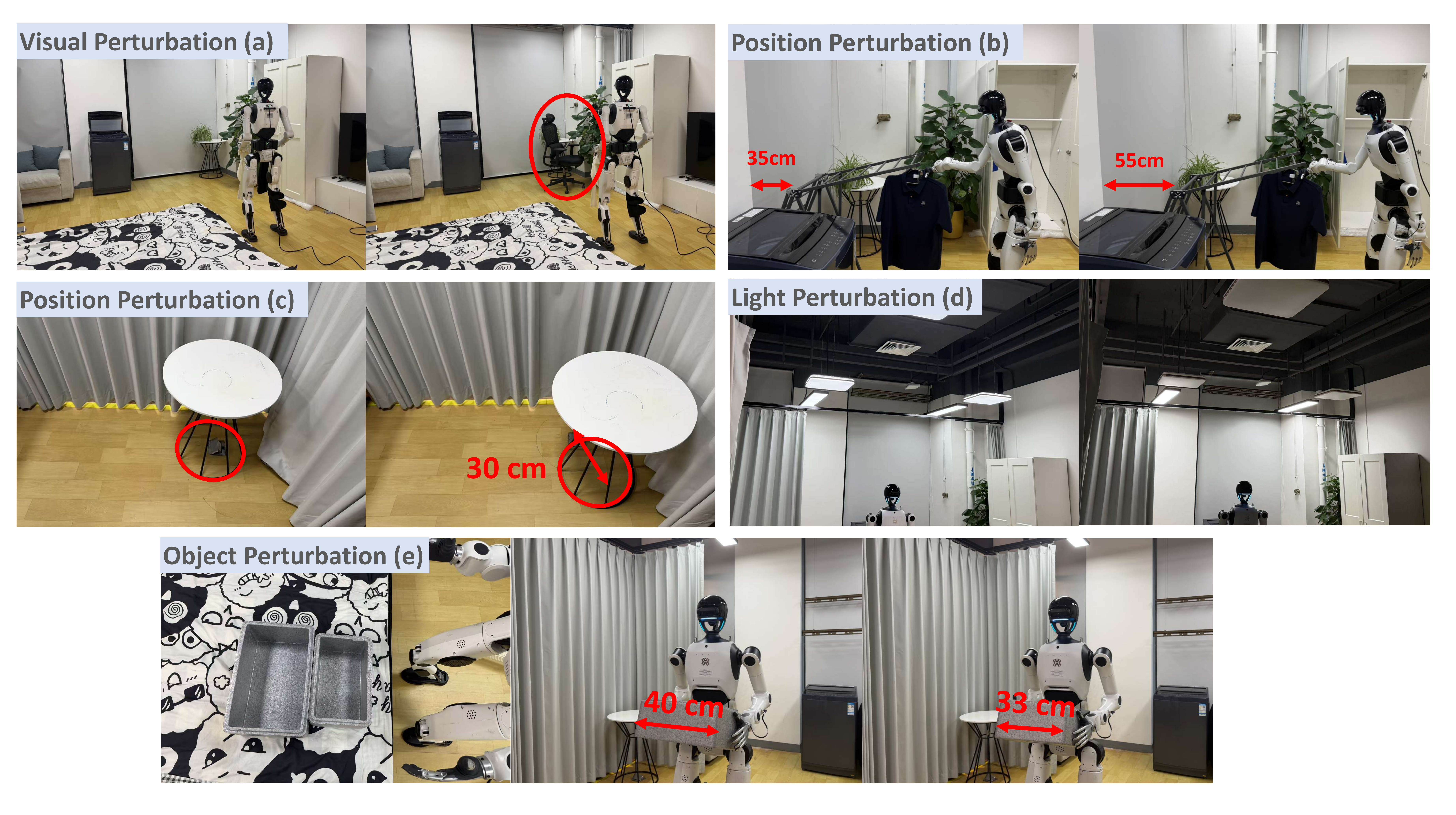}
  \caption{
\textbf{Generalization experiments under distribution shifts.}
(a) Unseen chair, (b) 20 cm initial-position shift,
(c) 30 cm target-table shift, (d) dimmed lighting,
and (e) smaller object. For each case, the unperturbed setting is shown on the left,
with the corresponding distribution-shifted setting shown on the right.
  }
  \label{fig:generalization}
  \vspace{-1em}
\end{figure}

We evaluate \ours under five controlled distribution shifts, as shown in \Cref{fig:generalization}.
For \textit{Laundry Loading}, we introduce an unseen office chair near the navigation path; for \textit{Clothes Retrieval}, we move the robot's initial position 20\,cm backward from the data-collection position.
For \textit{Box Carry}, we test dimmed lighting, a smaller manipulated object
($40\,\mathrm{cm}\times25\,\mathrm{cm}
\rightarrow33\,\mathrm{cm}\times20\,\mathrm{cm}$),
and a 30\,cm shift of the target table.
Together, these settings perturb visual appearance, object geometry, and spatial configuration.
\vspace{-0.3em}

As shown in \Cref{tab:generalization}, \ours achieves a higher normalized score than $\pi_{0.5}$ under all five perturbations and exhibits a smaller performance drop in four of them.
The advantage becomes particularly pronounced under spatial and geometric shifts.
For example, with a 30\,cm target-table shift, \ours retains 73.3\% compared with 40.0\% for $\pi_{0.5}$, corresponding to drops of 13.3 and 30.0 percentage points, respectively.
These results demonstrate stronger robustness beyond the original demonstration distribution.

\begin{table}[!htbp]
    \centering
    \small
    \setlength{\tabcolsep}{3.6pt}
    \renewcommand{\arraystretch}{1.12}
    \caption{
    \textbf{Generalization under controlled distribution shifts.}
    We report the average normalized task score (\%) and its absolute drop
    (\(\downarrow\), percentage points) from the corresponding unperturbed setting.
    }
    \label{tab:generalization}

    \begin{tabular}{cccccc}
        \toprule
        \multirow{2}{*}{\textbf{Task}}
        & \multirow{2}{*}{\textbf{Setting}}
        & \multicolumn{2}{c}{\textbf{HAF-VLA (Ours)}}
        & \multicolumn{2}{c}{$\boldsymbol{\pi_{0.5}}$} \\
        \cmidrule(lr){3-4}
        \cmidrule(lr){5-6}
        & &
        \textbf{Score}
        & \textbf{Drop}
        & \textbf{Score}
        & \textbf{Drop} \\
        \midrule

        Laundry Loading
        & Unseen Chair
        & \textbf{40.0}
        & $\downarrow$26.7
        & 26.7
        & $\downarrow$26.7 \\

        Clothes Retrieval
        & 20\,cm Start Shift
        & \textbf{43.3}
        & $\downarrow$10.0
        & 36.7
        & $\downarrow$16.7 \\

        \midrule

        \multirow{3}{*}{Box Carry}
        & Dim Lighting
        & \textbf{83.3}
        & $\downarrow$3.3
        & 60.0
        & $\downarrow$10.0 \\

        & Smaller Object
        & \textbf{76.7}
        & $\downarrow$10.0
        & 50.0
        & $\downarrow$20.0 \\

        & 30\,cm Table Shift
        & \textbf{73.3}
        & $\downarrow$13.3
        & 40.0
        & $\downarrow$30.0 \\

        \bottomrule
    \end{tabular}
\end{table}
\vspace{-1em}

\section{Limitations and Conclusion}
\label{sec:conclusion}

We present HAF-VLA, a hierarchical action-flow framework for adapting pretrained VLAs to humanoid whole-body loco-manipulation. By progressively denoising a shared action space and propagating intermediate motion plans through cross-stage KV-cache conditioning, HAF-VLA provides structured whole-body action generation for humanoid loco-manipulation. However, the hierarchical generation process increases the total denoising budget compared with a single-stage policy, which may introduce additional deployment latency. Moreover, our method still depends on the semantic reasoning and action priors of the underlying foundation model, and may fail when these priors are insufficient for unseen tasks or environments. Future work will explore more efficient denoising and deployment strategies to improve real-time performance, while further strengthening the foundation model adaptation for broader humanoid tasks.

\clearpage
\acknowledgments{This work was supported by the National Natural Science Foundation of China (62476011) and the Beijing Natural Science Foundation (L252060). We sincerely thank Jiaming Liu for his insightful ideas and valuable suggestions that helped shape this work. We are deeply grateful to Bowei Zhang for his great contributions to our experiments. We also appreciate our colleagues at the Beijing Innovation Center of Humanoid Robotics for their support in robot control, data collection, and technical assistance.}

\bibliography{references}

\clearpage
\appendix

\section*{Appendix}

\setcounter{section}{0}
\renewcommand{\thesection}{S\arabic{section}}
\renewcommand{\thesubsection}{\thesection.\arabic{subsection}}
\renewcommand{\thesubsubsection}{\thesubsection.\arabic{subsubsection}}

\setcounter{figure}{0}
\renewcommand{\thefigure}{S\arabic{figure}}

\setcounter{table}{0}
\renewcommand{\thetable}{S\arabic{table}}

\setcounter{equation}{0}
\renewcommand{\theequation}{S\arabic{equation}}

\section{Robot Action Interface}

\begin{figure}[htbp]
    \centering
    \includegraphics[width=0.8\textwidth]{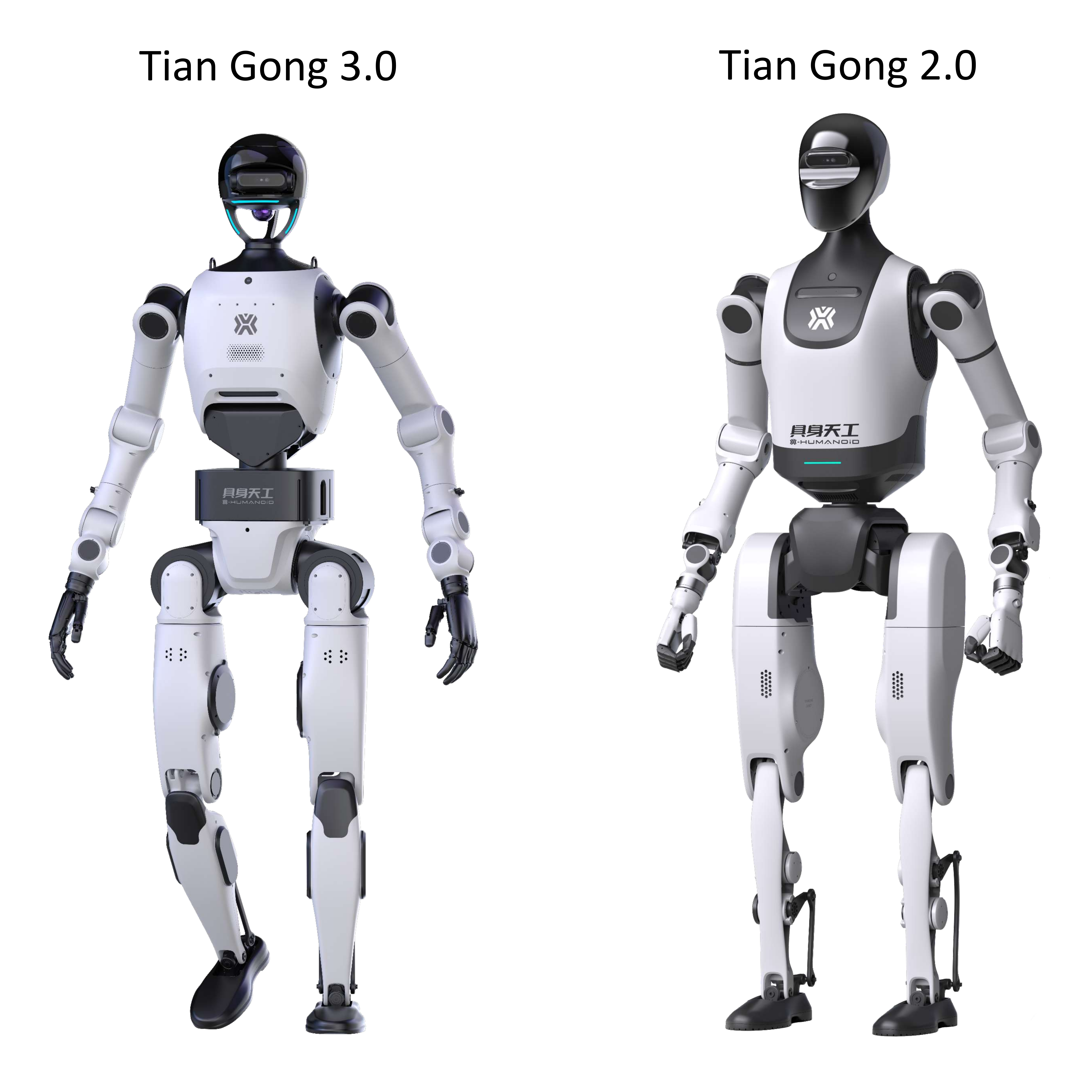}
    \caption{\textbf{Robot Platform.} We use Tian Gong 3.0 and Tian Gong 2.0 humanoid robots as our robot platforms.}
    \label{fig:robot_platform}
\end{figure}

We use Tian Gong 3.0 and Tian Gong 2.0 as our humanoid
robot platforms, as shown in \Cref{fig:robot_platform}. We use a unified policy-facing action interface for both humanoid platforms.
Although the internal robot control stacks are platform-dependent, the policy
always predicts actions in the same ordered action space:
\begin{equation}
\begin{aligned}
a_t^{ctrl} =
\big[
& q_{\mathrm{arm}}^{L},
  q_{\mathrm{arm}}^{R},
  g^{L},
  g^{R},
  v_x,
  v_y,
  \omega_z,
  q_{\mathrm{head}},
  w_{\mathrm{rpyz}},
  c_{\mathrm{skill}}
\big].
\end{aligned}
\end{equation}
The dimensions of these action components are defined as follows:
\begin{equation}
\begin{aligned}
q_{\mathrm{arm}}^{L}, q_{\mathrm{arm}}^{R} &\in \mathbb{R}^{7}, \\
g^{L}, g^{R} &\in [0,1], \\
(v_x, v_y, \omega_z) &\in \mathbb{R}^{3}, \\
q_{\mathrm{head}} &\in \mathbb{R}^{2}, \\
w_{\mathrm{rpyz}} &\in \mathbb{R}^{4}, \\
c_{\mathrm{skill}} &\in \{0,1,2\}.
\end{aligned}
\end{equation}

The two arm commands are target joint positions for the left and right arms,
and the two gripper commands \(g^{L}\) and \(g^{R}\) represent normalized
open-close commands. The locomotion command is represented by the planar
velocity command \((v_x, v_y, \omega_z)\). The vector \(w_{\mathrm{rpyz}}\)
denotes the posture-related command, whose interpretation depends on the target
platform. For Tian Gong 2.0, \(w_{\mathrm{rpyz}}\) specifies the desired
whole-body waist pose, which is realized through coordinated leg and waist
control. For Tian Gong 3.0, \(w_{\mathrm{rpyz}}\) corresponds to the three waist
joint targets and a squat-height command. The discrete skill command is defined
as
\begin{equation}
c_{\mathrm{skill}} =
\begin{cases}
0, & \text{walk}, \\
1, & \text{stand}, \\
2, & \text{kneel}.
\end{cases}
\end{equation}
Although the policy predicts the skill command as a continuous scalar, it is converted into a valid discrete skill index by the deployment wrapper after action denormalization. The resulting skill index is used to switch among the walking, standing, and kneeling control modes.

All action dimensions are normalized using quantile statistics computed from
the teleoperated demonstrations~\cite{intelligence2025pi05visionlanguageactionmodelopenworld}. Let \(q_{01}^{(d)}\) and \(q_{99}^{(d)}\)
denote the 1\% and 99\% quantiles of the \(d\)-th action dimension. During
training and deployment, we use the same quantile-based normalization and
inverse-normalization functions:
\begin{equation}
\tilde{a}_t^{(d)}
=
\mathcal{N}_{01/99}
\left(
a_t^{(d)}
\right),
\end{equation}
\begin{equation}
a_t^{(d)}
=
\mathcal{N}_{01/99}^{-1}
\left(
\tilde{a}_t^{(d)}
\right).
\end{equation}
This normalization reduces the influence of outliers in human teleoperation
data and keeps heterogeneous commands, such as joint targets, gripper commands,
velocity commands, and skill commands, in a consistent numerical range.

The lower-body and waist-related commands are executed by robot-specific
RL-based low-level controllers. The walking and waist-control behaviors are
trained using an AMP-style~\cite{amp} framework, while the kneeling behavior is trained
using a BeyondMimic-style~\cite{liao2025beyondmimicmotiontrackingversatile} motion-tracking controller. Therefore, HAF-VLA does
not directly output motor torques. Instead, it predicts high-level locomotion,
posture, and skill-mode commands, which are converted by the deployment wrapper
into the corresponding low-level controller inputs. The transition among
walking, standing, and kneeling is determined by \(c_{\mathrm{skill}}\). In
real-robot deployment, when switching from walking to standing, we insert an
approximately two-second zero-velocity safety waiting period before entering the
standing mode, which reduces abrupt mode transitions.

The upper-body commands are executed through position-control interfaces. The
arm commands are sent as target joint positions, and the gripper commands are
sent as normalized open-close values. The head command is represented as a
two-dimensional command in the shared policy interface. Tian Gong 3.0 actuates
two head degrees of freedom, while Tian Gong 2.0 only actuates the pitch degree
of freedom. For Tian Gong 2.0, the unused head dimension is ignored by the
platform-specific command wrapper. This abstraction allows the same policy-level
action representation to be used across both humanoid platforms, while
platform-dependent differences are handled only at deployment time.

We also use platform-specific wrappers to construct proprioceptive observations.

For Tian Gong 2.0, the proprioceptive state is
constructed as
\begin{equation}
\begin{aligned}
q_t^{\mathrm{TG2}} =
\big[
& q_{\mathrm{arm}}^{L},
  h^{L},
  q_{\mathrm{arm}}^{R},
  h^{R},
  w_{\mathrm{pose}},
  v_{\mathrm{leg}}
\big]
\in \mathbb{R}^{36}.
\end{aligned}
\end{equation}

The state components are defined as follows:
\begin{itemize}
    \item $q_{\mathrm{arm}}^{L}, q_{\mathrm{arm}}^{R}
    \in \mathbb{R}^{7}$:
    left and right arm joint positions.

    \item $h^{L}, h^{R} \in \mathbb{R}^{6}$:
    left and right hand states.

    \item $w_{\mathrm{pose}} \in \mathbb{R}^{4}$:
    waist pose.

    \item $v_{\mathrm{leg}} \in \mathbb{R}^{6}$:
    leg motion velocities.
\end{itemize}

For Tian Gong 3.0, the proprioceptive state is
constructed as
\begin{equation}
\begin{aligned}
q_t^{\mathrm{TG3}} =
\big[
& q_{\mathrm{arm}}^{L},
  q_{\mathrm{arm}}^{R},
  h^{L},
  h^{R},
  s_{\mathrm{touch}}^{L},
  s_{\mathrm{touch}}^{R},
  q_{\mathrm{head}},
  q_{\mathrm{leg}}^{L},
  q_{\mathrm{leg}}^{R},
  s_{\mathrm{stand}},
  x_{\mathrm{imu}}
\big]
\in \mathbb{R}^{99}.
\end{aligned}
\end{equation}

The state components are defined as follows:
\begin{itemize}
    \item $q_{\mathrm{arm}}^{L}, q_{\mathrm{arm}}^{R}
    \in \mathbb{R}^{7}$:
    left and right arm joint positions.

    \item $h^{L}, h^{R} \in \mathbb{R}^{6}$:
    left and right hand states.

    \item $s_{\mathrm{touch}}^{L}, s_{\mathrm{touch}}^{R}
    \in \mathbb{R}^{20}$:
    left and right tactile sensor states.

    \item $q_{\mathrm{head}} \in \mathbb{R}^{2}$:
    head joint positions.

    \item $q_{\mathrm{leg}}^{L}, q_{\mathrm{leg}}^{R}
    \in \mathbb{R}^{6}$:
    left and right leg joint positions.

    \item $s_{\mathrm{stand}} \in \mathbb{R}^{6}$:
    standing-status representation.

    \item $x_{\mathrm{imu}} \in \mathbb{R}^{13}$:
    selected body IMU features used by the policy.
\end{itemize}

The proprioceptive states are normalized using the same percentile-based
strategy before being fed into the policy.
\section{Implementation Details}

\subsection{HAF-VLA Implementation Details}

HAF-VLA is implemented on top of the pretrained $\pi_{0.5}$ model following the official openpi codebase~\cite{intelligence2025pi05visionlanguageactionmodelopenworld}. We initialize the vision-language prefix encoder and the action expert from the pretrained $\pi_{0.5}$ checkpoint, and add learnable stage embeddings, stage-wise action masks, cross-stage KV-cache conditioning, and the 
action input/output projection layers for hierarchical whole-body action generation. The action horizon is set to 100. During inference, HAF-VLA performs three sequential denoising stages with 10 flow-matching steps per stage. The final action chunk is taken from Stage 3, and the robot executes the first 40 steps before the next inference call.

For parameter-efficient adaptation, we use the \texttt{gemma\_2b\_lora} variant in openpi and apply LoRA~\cite{hu2021loralowrankadaptationlarge} to the PaliGemma~\cite{beyer2024paligemmaversatile3bvlm} visual language backbone. The LoRA rank is 16, the LoRA alpha is 16.0, and the resulting scaling factor is 1.0. We do not use LoRA dropout. Following the official implementation, LoRA is applied to both attention and feed-forward layers, including the query projection, key-value projection, output projection, gated/up projection, and down projection.

The original weights of the PaliGemma language backbone are frozen except for the LoRA parameters. The action expert uses the \texttt{gemma\_300m} variant and remains trainable. The newly introduced hierarchy-specific parameters, including stage embeddings, action input/output projections, and time MLP layers, are also trainable. The vision encoder is not frozen in our current configuration.

We train HAF-VLA with AdamW~\cite{loshchilov2019decoupledweightdecayregularization} using $\beta_1=0.9$, $\beta_2=0.95$, $\epsilon=10^{-8}$, weight decay $10^{-10}$, and global gradient norm clipping of 1.0. The learning rate follows a cosine decay schedule with 1000 warmup steps, a peak learning rate of $2.5\times10^{-5}$, a decay length of 30k steps, and a final learning rate of $2.5\times10^{-6}$. Unless otherwise specified, the model is trained for 50k steps with a global batch size of 64.

 For state input, since we use the $\pi_{0.5}$ configuration with discrete state input, the robot state is discretized and appended to the prompt tokens, with a maximum token length of 385 for Tian Gong 3.0 and 192 for Tian Gong 2.0.

\subsection{Baseline Implementation Details}
\label{sec:supp_baseline_impl}

For all baseline methods, we use the same teleoperated training data, humanoid action interface, and robot-specific deployment wrapper as HAF-VLA. The predicted action chunks are mapped to the same low-level locomotion, waist, head, arm, and hand control interfaces for fair comparison.

\paragraph{$\pi_{0.5}$~\cite{intelligence2025pi05visionlanguageactionmodelopenworld}.}
We use $\pi_{0.5}$ as a strong single-stage VLA baseline. $\pi_{0.5}$ is a large-scale general-purpose robot foundation model in which the vision-language model and the action expert share the same attention backbone. We fine-tune the full model on our humanoid demonstration data for better performance. The action horizon is set to 100, and the model is trained for 50k steps on 8 NVIDIA A100 GPUs with a per-GPU batch size of 8. During deployment, we execute the first 40 steps of each predicted action chunk before the next inference call.

\paragraph{ACT~\cite{act}.}
ACT is used as a compact imitation-learning baseline with a Transformer encoder-decoder architecture and action chunking. We set the action horizon to 200 and train the model for 37k steps on 8 NVIDIA A100 GPUs with a per-GPU batch size of 64. During deployment, the model outputs action chunks with temporal aggregation over a 60-step window, which smooths the final commands sent to the robot.

\paragraph{GR00T N1.7~\cite{gr00tn1_2025}.}
GR00T N1.7 is used as a humanoid foundation-model baseline. It predicts continuous action sequences with a vision-language backbone and a diffusion Transformer action head. We set the action horizon to 100 and fine-tune the model for 20k steps on 8 NVIDIA A100 GPUs with a global batch size of 160. During deployment, we use an action chunk size of 50, which achieves the best empirical performance in our experiments.

\paragraph{Cosmos Policy~\cite{kim2026cosmos}.}
Cosmos Policy is used as a world-model-based visuomotor-control baseline. It is built on the Cosmos framework and uses the Cosmos-Predict2-2B backbone together with a diffusion Transformer action head to predict continuous robot action sequences. We set the action horizon to 50 and train the policy for 15k steps on 8 NVIDIA A100 GPUs with a global batch size of 160. During deployment, we use an action chunk size of 40, which gives the best empirical performance in our experiments.

\subsection{Implementation Details of Q2 Ablation Variants}

\paragraph{Notation and common setup.}
Following the notation in the main paper, the model action dimension is
$D=32$, comprising 26 effective control dimensions and 6 zero-padded
dimensions. We partition the effective dimensions into three disjoint groups:
$\mathcal{G}_{L}
= \mathcal{D}_{\mathrm{move}}
\cup \mathcal{D}_{\mathrm{skill}}
\cup \mathcal{D}_{\mathrm{head}}$,
$\mathcal{G}_{W}=\mathcal{D}_{\mathrm{waist}}$, and
$\mathcal{G}_{A}
= \mathcal{D}_{\mathrm{arm}}
\cup \mathcal{D}_{\mathrm{hand}}$.
The proposed hierarchy uses the cumulative active dimension sets
\begin{equation}
    \mathcal{D}_1 = \mathcal{G}_{L}, \qquad
    \mathcal{D}_2 = \mathcal{G}_{L}\cup\mathcal{G}_{W}, \qquad
    \mathcal{D}_3 = \mathcal{G}_{L}\cup\mathcal{G}_{W}
    \cup\mathcal{G}_{A}=\mathcal{D}.
\end{equation}
Let $m_s\in\{0,1\}^{D}$ denote the binary action-dimension mask of stage $s$,
broadcast along the temporal dimension. The padding dimensions remain
inactive at all stages.

Following the flow-matching formulation in the main paper,
given a zero-padded clean action chunk $A_t\in\mathbb{R}^{H\times D}$,
Gaussian noise $\epsilon\sim\mathcal{N}(0,I)$,
and flow time $\tau\in[0,1]$,
the noisy input and target velocity field of stage $s$ are
\begin{equation}
    X_\tau^s
    =\bigl((1-\tau)\epsilon+\tau A_t\bigr)\odot m_s,
    \qquad
    u_s=(A_t-\epsilon)\odot m_s.
\end{equation}
Inactive dimensions are zeroed at the input and assigned
zero target velocities.
Unless otherwise specified, the stage-wise flow-matching loss is
\begin{equation}
    \mathcal{L}_s
    =\frac{1}{BHD}
    \sum_{b,h,d}
    \left(
    \hat u_{s,b,h,d}-u_{s,b,h,d}
    \right)^2,
    \qquad
    \mathcal{L}_{\mathrm{FM}}
    =\sum_{s=1}^{3}\mathcal{L}_s,
\end{equation}
where $B$ and $H$ denote the batch size and action horizon.
The loss is averaged over all $D=32$ action dimensions,
including inactive and padded dimensions whose target velocities are zero.

All stages share the same action expert while retaining their
respective stage embeddings.
Each stage attends to the shared vision-language cache
$C_{\mathrm{vl}}$, its current action tokens, and the available
preceding action caches $C_{<s}$.
During training, cross-stage caches are constructed by
re-encoding the corresponding masked ground-truth action chunks.
These caches are detached before being passed to subsequent stages.
During inference, they are constructed from the completed
denoised predictions of preceding stages.
Unless otherwise specified, each stage performs 10 Euler
denoising steps, and the final executed action chunk is
obtained from Stage 3.

\paragraph{Stage-order variants.}
We modify the order in which action groups are introduced
while preserving the cumulative three-stage structure.
\textit{Waist-First} uses
$\mathcal{G}_{W}\rightarrow\mathcal{G}_{A}
\rightarrow\mathcal{G}_{L}$,
whereas \textit{Arm-First} uses
$\mathcal{G}_{A}\rightarrow\mathcal{G}_{W}
\rightarrow\mathcal{G}_{L}$.
For both variants, each stage activates all groups introduced
up to that stage, and Stage 3 activates all effective action dimensions.
The original action-vector layout is preserved; only the
stage masks are modified.
Both variants retain the shared action expert, stage embeddings,
cross-stage cache conditioning, and loss aggregation.
As in the proposed hierarchy, later stages may refine
action dimensions introduced by earlier stages.

\paragraph{Loss-Calibrated.}
This variant preserves the proposed stage ordering and
modifies only the elementwise training loss.
For each effective action dimension $d$, we define
\begin{equation}
    c_d=\sum_{s=1}^{3}m_s^{(d)},
\end{equation}
which counts the number of stages in which that dimension
is active.
We apply inverse-frequency weighting to active dimensions:
\begin{equation}
    w_{s,d}=
    \begin{cases}
        1/c_d, & m_s^{(d)}=1,\\
        1, & m_s^{(d)}=0,
    \end{cases}
\end{equation}
and compute
\begin{equation}
    \mathcal{L}_s^{\mathrm{cal}}
    =\frac{1}{BHD}
    \sum_{b,h,d}
    w_{s,d}
    \left(
    \hat u_{s,b,h,d}-u_{s,b,h,d}
    \right)^2.
\end{equation}
Accordingly, active locomotion/head/skill, waist, and
arms/hands dimensions receive weights of $1/3$, $1/2$,
and $1$, respectively.
Inactive and padded dimensions retain unit-weight
zero-target supervision.
The normalization denominator remains $BHD$
rather than the sum of elementwise weights.
Inference is identical to the proposed method.

\paragraph{All-Joint.}
This variant removes body-part-specific staging by activating
all effective action dimensions at every stage:
\begin{equation}
    \mathcal{D}_1
    =\mathcal{D}_2
    =\mathcal{D}_3
    =\mathcal{D}.
\end{equation}
Consequently, every stage generates a complete effective action chunk,
while the six padded dimensions remain inactive.
The shared action expert, stage embeddings, and cross-stage
cache conditioning are preserved.
During inference, each stage starts from independently
sampled Gaussian noise and conditions on the available
preceding action caches.
Only the final Stage 3 prediction is executed.

\paragraph{Denoising-budget control.}
To distinguish hierarchical generation from simply increasing
the denoising budget, we compare the proposed model,
which performs 10 Euler denoising steps at each of its
three stages, with single-stage $\pi_{0.5}$ using 30 steps.
The latter generates the complete action chunk through
a single denoising trajectory without intermediate
action-stage conditioning.
This comparison matches the total number of denoising
updates but not necessarily the computational cost,
since the hierarchical model additionally constructs
cross-stage action caches and uses different
attention-context lengths across stages.

\paragraph{No-Cache-Train.}
We disable cross-stage action-cache construction and
conditioning during both training and inference.
Each stage conditions only on the shared vision-language
cache $C_{\mathrm{vl}}$ and its own current action tokens.
The stage embeddings, cumulative action masks,
zero-target supervision, and stage-wise losses are unchanged.
During inference, the stages use independent noise
initializations, and the final Stage 3 prediction
provides the executed action chunk.

\paragraph{No-Cache-Deploy.}
We retain the normally trained checkpoint but disable
attention to preceding action-stage caches during inference.
Specifically, the attention mask prevents Stage 2 from
attending to $C_1$ and Stage 3 from attending to
$C_1$ and $C_2$.
The cached tensors, sequence lengths, and positional
indices are preserved, while the vision-language cache
and current-stage tokens remain accessible.
Thus, this variant removes access to preceding action
representations without changing the attention-token layout.

\paragraph{Zero-Cache-Deploy.}
Using the same normally trained checkpoint, we instead
replace the keys and values of all preceding action-stage
cache tokens with zeros at every transformer layer.
The vision-language cache remains unchanged.
Unlike \textit{No-Cache-Deploy}, the zeroed action-cache
tokens remain visible to attention and participate
in softmax normalization.
The two variants therefore introduce distinct
inference-time disruptions to cross-stage conditioning.

\paragraph{Hard-Freeze.}
This variant retains cross-stage cache conditioning but
prevents later stages from modifying action dimensions
introduced by earlier stages.
We define the inherited-dimension mask and the
newly introduced-dimension mask as
\begin{equation}
    i_s=\bigvee_{r<s}m_r,
    \qquad
    n_s=m_s\odot(1-i_s),
\end{equation}
where $i_1=0$.
During training, inherited dimensions are fixed to
their clean ground-truth values, while only newly
introduced dimensions undergo flow matching:
\begin{equation}
    X_\tau^s
    =i_s\odot A_t
    +n_s\odot
    \bigl((1-\tau)\epsilon+\tau A_t\bigr),
    \qquad
    u_s=n_s\odot(A_t-\epsilon).
\end{equation}
The loss excludes inherited dimensions but retains
zero-target supervision for inactive dimensions:
\begin{equation}
    \mathcal{L}_s^{\mathrm{freeze}}
    =\frac{1}{BHD}
    \sum_{b,h,d}
    (1-i_s^{(d)})
    \left(
    \hat u_{s,b,h,d}-u_{s,b,h,d}
    \right)^2.
\end{equation}

During inference, inherited values are taken from
the completed prediction of the preceding stage,
while only newly introduced dimensions are updated
through Euler denoising.
After every update, inherited values are restored
and inactive dimensions are zeroed.
Consequently, the final action chunk strictly preserves
the locomotion/head/skill prediction from Stage 1
and the waist prediction introduced at Stage 2.
As in the proposed method, training uses clean
ground-truth actions for cross-stage conditioning,
whereas inference uses model predictions.

\paragraph{Stage-Concat.}
This variant preserves the normal three-stage generation
and cache-conditioning procedure but changes how the
final action chunk is assembled for execution.
Instead of executing the jointly refined Stage 3 output,
we concatenate the action groups from their respective
introduction stages:
\begin{equation}
    A_{\mathrm{exec}}
    =m_{\mathcal{G}_{L}}\odot\hat A_1
    +m_{\mathcal{G}_{W}}\odot\hat A_2
    +m_{\mathcal{G}_{A}}\odot\hat A_3,
\end{equation}
where $m_{\mathcal{G}_{L}}$,
$m_{\mathcal{G}_{W}}$, and
$m_{\mathcal{G}_{A}}$ denote the fixed masks
of the corresponding action groups.
Each selected action group retains its original
position in the full action vector.

Unlike \textit{Hard-Freeze}, later stages are still
allowed to refine earlier action dimensions internally.
However, these refinements are discarded when constructing
the executed action.
This variant therefore isolates the contribution of
executing the final jointly refined action prediction
without modifying the underlying hierarchical generation.
\section{Detailed Task Description}
\label{sec:supp_task_description}

\paragraph{Evaluation protocol.}
We evaluate HAF-VLA and the baseline methods on six real-world
humanoid loco-manipulation tasks in the main benchmark (Q1).
An additional task, \textit{Box Carry}, is used for the
controlled ablation studies in Q2. All seven tasks follow
the same milestone-based evaluation protocol.
Each rollout starts from a predefined region that matches
the distribution used during data collection.
For fair comparison, different methods are evaluated using
the same set of initial configurations whenever possible.
Since these tasks involve long-horizon execution and multiple
intermediate objectives, we use milestone-based normalized
task scores instead of binary success rates.
Each task has a maximum score of 15, and the final normalized
score is computed as the achieved milestone score divided
by the maximum score.

\paragraph{Language instructions.}
For each task, the policy receives a task-level language instruction together with egocentric visual observations and proprioceptive robot states. The language instructions used in our experiments are:
\begin{itemize}
\item \textit{Laundry Loading}: ``Pick up the clothes on the bed, walk to the washing machine, and put the clothes into the washing machine.''
\item \textit{Clothes Retrieval}: ``Pick up the clothes from the drying rod, walk to the wardrobe and hang the clothes in the wardrobe.''
\item \textit{Table Tidy}: ``Walk to the table, close the laptop, grasp the bottle, and throw it into the trash bin.''
\item \textit{Plant Watering}: ``Walk to the shelf, lift up the flower pot with the right hand, and water the flowers.''
\item \textit{Box Transfer}: ``Carry the box, turn around, walk to the table, and put it down.''
\item \textit{Ball Tossing}: ``Grasp the ball, walk to the table, and drop the ball into the box.''
\item \textit{Box Carry}: ``Carry the box to the table.''
\end{itemize}

\paragraph{Laundry Loading.}
In the \textit{Laundry Loading} task, the robot starts near the bed, reaches toward scattered clothes, grasps the clothes from the bed surface, walks to the washing machine, places the clothes into the washing machine, and finally closes the washing-machine lid. During data collection, the robot starts from three positions around the bed: approximately 30 cm to the left of the center, at the center, and approximately 30 cm to the right of the center. This task requires the robot to combine flexible-object grasping, spatial navigation between the bed and the washing machine, and final object placement into a constrained container. The major challenge comes from the deformable nature of clothes: their shape, contact points, and grasp affordances vary significantly across rollouts, making robust grasping and stable carrying difficult. The robot must also accurately align with the washing-machine opening before releasing the clothes. The maximum score is 15: placing one piece of clothing into the washing machine gives 5 points, placing two pieces gives 10 points, and successfully closing the lid gives an additional 5 points.

\paragraph{Clothes Retrieval.}
In the \textit{Clothes Retrieval} task, the robot approaches a drying rod, reaches upward to remove a piece of clothing, carries it to the wardrobe, hangs or places it inside the wardrobe, steps backward, and closes the wardrobe door. During data collection, the clothing is placed at four uniformly spaced positions along the drying rod, covering different relative positions between the robot and the target clothing. This task requires high-position object interaction, dual-arm coordination, and precise whole-body positioning. A key challenge is that the robot must stand at an appropriate distance from the drying rod and the wardrobe: if it stops too far away, the arms must overextend; if it stops too close, the arms and torso may collide with the environment. The task also requires the robot to coordinate backward motion and arm movement when closing the wardrobe door. The maximum score is 15: successfully removing the clothes from the drying rod gives 5 points, placing or hanging the clothes in the wardrobe gives 5 points, and closing the wardrobe door gives an additional 5 points.

\paragraph{Table Tidy.}
In the \textit{Table Tidy} task, the robot walks to a table, approaches the laptop, closes the laptop lid, identifies and grasps a bottle or piece of trash on the tabletop, and throws it into the trash bin. During data collection, the robot starts from six positions in front of the table: two rows at different distances from the table, with three lateral offsets in each row, approximately 15 cm left, centered, and 15 cm right. This task contains multiple heterogeneous subgoals within a single rollout. The robot must first perform a contact-rich manipulation action to close the laptop, then switch to grasping and throwing another object. The main challenges are precise tabletop alignment, object-level spatial understanding, and robust transition between different manipulation modes. Small base-position errors can make the laptop-closing action miss the lid or cause the trash-grasping motion to be misaligned. The maximum score is 15: closing the laptop gives 5 points, successfully grasping the trash gives 5 points, and throwing it into the trash bin gives an additional 5 points.

\paragraph{Plant Watering.}
In the \textit{Plant Watering} task, the robot moves toward a shelf with plants at different heights, lifts and stabilizes the flower pot or watering container with the right hand, waters the upper plant, changes its body posture through kneeling or lowering the waist, and then waters the lower plant. During data collection, the initial robot position is randomly perturbed within a small range. This task emphasizes whole-body posture adaptation. The robot must coordinate locomotion, waist motion, arm motion, and skill-mode control to reach plants at different heights. The main challenge is that the robot cannot solve the task using arm motion alone: it must adjust its body height and posture to reach the lower target while maintaining balance and keeping the container oriented correctly. The maximum score is 15: successfully watering the upper plant gives 5 points, successfully entering the kneeling or lowered-body posture gives 5 points, and successfully watering the lower plant gives an additional 5 points.

\paragraph{Box Transfer.}
In the \textit{Box Transfer} task, the robot grasps a box from one table, lifts and carries it, turns around, walks to another table, re-aligns its body with the target table, and places the box down. During data collection, the relative positions of the robot and the source table are perturbed within a small range. This task requires stable object carrying during locomotion and accurate final placement after navigation. The main challenge is that the robot must maintain the box pose while walking and turning; otherwise, the box may slip, tilt, or collide with the robot body. After reaching the target table, the robot must also re-localize the placement surface and adjust its arm trajectory accordingly. The maximum score is 15: successfully lifting the box gives 5 points, moving to the target table while carrying the box gives 5 points, and successfully placing the box on the target table gives an additional 5 points.

\paragraph{Ball Tossing.}
In the \textit{Ball Tossing} task, the robot reaches toward a ball on the table, grasps it, moves toward the target box, aligns its body and arm with the box, and releases the ball into the box. During data collection, the ball is placed at nine annotated positions on the source table, and the target box is also placed at nine annotated positions; the paired ball and box positions are collected sequentially to cover different grasping and releasing geometries. This task is particularly sensitive to final-stage execution accuracy. In addition to grasping and locomotion, the robot must control the release timing, wrist pose, and body alignment so that the ball enters the box. Even small errors in base position or release direction may cause the ball to miss the box. The maximum score is 15: successfully grasping the ball gives 5 points; if the released ball does not enter the box but touches the box wall, the rollout receives 5 points for partial release accuracy; if the ball successfully enters the box, the release stage receives 10 points.

\paragraph{Box Carry.}
In the \textit{Box Carry} task, the robot picks up a box from a bed, carries it while walking toward a round table, and places it on the tabletop. This task evaluates the robot's ability to coordinate whole-body locomotion and manipulation while transporting an object between different surfaces. The robot must maintain a stable grasp and keep the box balanced throughout locomotion, avoiding unintended contact with its surroundings. Upon approaching the target table, it must stop safely without collisions and adjust its body posture to prepare for placement. The main challenge lies in maintaining object stability during locomotion while achieving precise placement on the relatively small, circular tabletop. The maximum score is 15: successfully lifting the box earns 5 points, reaching the target table and stopping without collisions earns 5 points, and placing the box entirely within the tabletop boundary earns an additional 5 points.
\subsection{Full-Task Success Rates and Statistical Uncertainty}

In addition to the normalized task scores reported in
\Cref{tab:vla_score}, we evaluate full-task success rates
to distinguish complete task execution from partial progress.
A rollout is considered successful only when the entire task
is completed.
For each method and task, we report the success rate with
a 95\% Wilson confidence interval, computed from 10 real-robot
rollouts.
The results are summarized in \Cref{tab:success_rate}.
Given the limited number of trials, the confidence intervals
remain relatively wide and should be interpreted as estimates
of statistical uncertainty rather than evidence of
statistically significant differences.

\begin{table}[htbp]
    \centering
    \footnotesize
    \setlength{\tabcolsep}{3.5pt}
    \renewcommand{\arraystretch}{1.15}
    \caption{
    \textbf{Full-task success rates with statistical uncertainty.}
    We report success rates (\%) with 95\% Wilson confidence
    intervals over 10 real-robot rollouts per method and task.
    Each entry is formatted as success rate [lower, upper].
    }
    \label{tab:success_rate}

    \begin{tabular}{lccccc}
        \toprule
        \textbf{Task}
        & \textbf{HAF-VLA (Ours)}
        & $\boldsymbol{\pi_{0.5}}$
        & \textbf{GR00T N1.7}
        & \textbf{Cosmos Policy}
        & \textbf{ACT} \\
        \midrule

        Laundry Loading
        & 30 [11, 60]
        & 20 [6, 51]
        & 10 [2, 40]
        & 0 [0, 28]
        & 10 [2, 40] \\

        Clothes Retrieval
        & 40 [17, 69]
        & 30 [11, 60]
        & 20 [6, 51]
        & 0 [0, 28]
        & 0 [0, 28] \\

        Table Tidy
        & 60 [31, 83]
        & 40 [17, 69]
        & 20 [6, 51]
        & 0 [0, 28]
        & 10 [2, 40] \\

        Plant Watering
        & 50 [24, 76]
        & 40 [17, 69]
        & 20 [6, 51]
        & 20 [6, 51]
        & 0 [0, 28] \\

        Ball Tossing
        & 40 [17, 69]
        & 20 [6, 51]
        & 20 [6, 51]
        & 0 [0, 28]
        & 10 [2, 40] \\

        Box Transfer
        & 80 [49, 94]
        & 30 [11, 60]
        & 20 [6, 51]
        & 30 [11, 60]
        & 30 [11, 60] \\

        \bottomrule
    \end{tabular}
\end{table}

\end{document}